\documentclass{article}
\usepackage{ijcai21}
\usepackage{float}
\usepackage{times}

\usepackage{soul}
\usepackage{url}
\usepackage[hidelinks]{hyperref}
\usepackage[utf8]{inputenc}
\usepackage[small]{caption}
\usepackage{graphicx}
\usepackage{amsmath}
\usepackage{booktabs}
\usepackage{mathtools}

\DeclarePairedDelimiter\abs{\lvert}{\rvert}%
\makeatletter
\let\oldabs\abs
\def\abs{\@ifstar{\oldabs}{\oldabs*}}

\usepackage{xcolor}

\title{Exploiting Spline Models for the Training of Fully Connected Layers in Neural Network}

\author{
Kanya Mo$^1$ \footnote{The first three authors have equal contribution.}\and
Shen Zheng$^1$\and
Xiwei Wang$^1$\and
Jinghua Wang$^2$\And
Klaus-Dieter Schewe$^1$\footnote{Contact Author}\\
\affiliations
$^1$Zhejiang University, UIUC Institute, Haining, China\\
$^2$University of Illinois at Urbana-Champaign, Urbana, IL, USA\\
\emails
\{Kanya.18, Shen.18, Xiwei.18\}@intl.zju.edu.cn,
jinghua3@illinois.edu,
kdschewe@acm.org
}

\begin{document}

\maketitle

\begin{abstract}
The fully connected (FC) layer, one of the most fundamental modules in artificial neural networks (ANN), is often considered difficult and inefficient to train due to issues including the risk of overfitting caused by its large amount of parameters. Based on previous work studying ANN from linear spline perspectives, we propose a spline-based approach that eases the difficulty of training FC layers. Given some dataset, we first obtain a {\it continuous piece-wise linear} (CPWL) fit through spline methods such as {\it multivariate adaptive regression spline} (MARS). Next, we construct an ANN model from the linear spline model and continue to train the ANN model on the dataset using gradient descent optimization algorithms. Our experimental results and theoretical analysis show that our approach reduces the computational cost, accelerates the convergence of FC layers, and significantly increases the interpretability of the resulting model (FC layers) compared with standard ANN training with random parameter initialization followed by gradient descent optimizations.
\end{abstract}

\section{Introduction}
Deep neural networks (DNN) have significantly increased the capability of dealing with difficult machine learning and signal processing problems. A DNN is composed of a large number of simple parameterized linear and nonlinear transformations. Previous work \cite{pmlr-v80-balestriero18b} has shown a rigorous bridge between deep networks (DNs) and spline operators, but the insights have not yet been exploited in applications. We also noticed that although FC is one of the fundamental components of standard and convolutional neural networks (CNN), it may be inefficient to train and also cause possible overfitting problems due to an abundance of parameters \cite{simonyan2015deep}. Therefore, more and more deep models prefer discarding FC for better performance and efficiency \cite{Deep-Residual,Network-In-Network,2015Bilinear,Mask-CNN,Mask-CNN}. However, a recent paper \cite{Zhang:In-defense} also defends for FC and reclaims it importance in visual representation transfer. Despite the redundant parameters and possible overfitting, FC layers serve as powerful feature extractors and classifiers. It is therefore of great significance to investigate effective and efficient ways for the training of FC layers.

In this paper, we propose an algorithm that first builds a continuous piece-wise linear (CPWL) fit through a linear spline method, then converts the fit to several FC layers and continues training. As the converted FC layers can be trained using gradient descent, our algorithm can be viewed as an initialization of parameters. In the following theoretical analysis, we also demonstrate that the  approach promotes the training speed and generally increases interpretability when viewed from the results of spline fit.  We further conducted experiments on {\it Abalone} dataset \cite{Abalone} and {\it Wine Quality} dataset \cite{Wine-quality}, which further underlines the practical feasibility of our algorithm.

\section{Background}

\subsection{Multivariate Adaptive Regression Spline (MARS)}
The {\it multivariate adaptive regression splines} (MARS) is a flexible regression model for large volumes of data with high dimensions \cite{friedman1991multivariate}. A MARS model is the linear combination of {\it spline basis functions} in the form of
\begin{equation}
    f(X) = \sum_{m=0}^{M} \beta_{i}h_{m}(X) 
\end{equation}
where the spline basis functions have the form
\begin{equation}
    h_m(X)=
    \begin{cases}
    1&\text{if}\; m=0\\
    \prod_{i=1}^{K_m}R(\pm(X_{mi}-t_{mi}))&\text{if}\; m > 0
    \end{cases}
\end{equation}
and $R(x)$ is a ramp function
\begin{equation}
    R(x)=
    \begin{cases}
    x& \text{x$\geq$0}\\
    0& \text{x$\textless$0}
    \end{cases}
\end{equation}

All $K_m$ is not larger than $K_{max}$, {\it the upper limit on the order of interaction}. The functions $R(X_{mi}-t_{mi})$ and $R(t_{mi}-X_{mi})$ are called as a {\it reflection pair}. To obtain a CPWL fit with MARS, $K_{max}$ is set to $1$ and MARS degenerates to a first-order additive model.

The spline basis functions and the parameters in a MARS model are determined by a step-wise forward adaptation process followed by a step-wise backward pruning process. The fitting criterion of MARS is to minimize {\it generalized cross-validation} (GCV)
\begin{equation}
    GCV(M)=\frac{\sum_{i=1}^{N}(y_i-\widehat{f}_{M}(x_i))^2}{(1-C(M)/N)^2}
\end{equation}
where $C(M)$ is the effective number of parameters being fit for fixed M; this is done in the same way as linear regression.

The step-wise forward adaptation process selects all functions $h_m(X)$ starting from $h_0(X)=1$ and adding a reflection pair each time. The new basis functions are the respective products of the reflection pair $R(\pm(X_{j}-t))$ and one of the selected basis functions. For first-order MARS, the new basis functions can only be the reflection pair $R(\pm(X_{j}-t))$ and the intercept. In each iteration, subject to the upper limit $K_{max}$, the input variable $X_{j}$, the {\it knot value} $t$ and the existing basis function are chosen such that $GCV(M)$ is minimized. Here, the {\it knot value} is chosen from the input values. The procedure stops when the number of basis functions reaches a preset upper limit $M_{max}$.

The step-wise backward pruning process discards some of the spline basis functions determined in the forward phase to avoid the overfitting problem. The pruning procedure gradually deletes an arbitrary basis function except $h_0(x)=1$ such that a new smaller model with minimized $GCV(M)$ results. The procedure stops when no further improvement is possible or the model only contains $h_0(x)=1$. During the backward phase starting with the original model, we obtain up to $M_{max}$ models of decreasing size, so we choose the one minimizing $GCV(M)$ as the final MARS model.

\subsection{Fully Connected Neural Network and Its Spline Perspective}

The {\it fully connected network} (FC network) is the neural network in which all neurons are connected to all the neurons in the neighbored layers. Such a neural network can be expressed as a composition of {\it FC operators} and {\it activation operators}. Here we define {\it FC operators} as follows:
\begin{equation}
    f(x) := W_{k}x + b_{k}
\end{equation}
where ${W_{k}}$ is the weight matrix in layer $k$, ${b_{k}}$ is the bias vector in layer $k$ and $x$ is the input vector or the output vector from the previous layer. The {\it activation operator} applies a (nonlinear) activation function on each input entry. 

Though neural networks have been long recognized as a black box model, some deep insights from a spline perspective have already been gained. Arora et al. have shown that any PWL function can be represented by a single-output ReLU DNN and that the converse of the argument is also true \cite{Arora-Rectified-Linear-Units}, which builds a strong bridge between ReLU networks and linear splines. In a more recent work \cite{pmlr-v80-balestriero18b}, it has been shown that a large class of DNs with piecewise-affine {\it activation operators} can be represented by max-affine spline operators. The max-affine spline operator is a multivariable extension of max-affine functions, which is of the form \cite{convex}:
\begin{equation}
    f(x) = \max_{i = 1...k} {a_{i}^{T}x + b_{i}}
\end{equation}


\section{Method}
In this section, we propose our method in three parts, motivations, a spline-based approach boosting the training of ReLU networks and theoretical extensions of our method.
\subsection{Motivation}
The background section mentions a spline view into neural networks with PWL activation functions. For networks with other types of {\it activation operators}, they can also be well approximated by PWL functions if we consider a network with the same shape and the {\it activation operator} replaced by its PWL approximation. For networks using certain types of activation functions that can be easily approximated by ReLU, such as the sigmoid function, it is also feasible to approximate the network with ReLU networks and therefore, linear splines.

Beside these connection between neural networks and splines, we also notice that it is overall convenient to obtain a PWL fit to a given set of sample points. There are methods in spline theory that calculate a linear combination of spline basis using either explicit formulae or recursive algorithms like MARS, and these methods are usually faster in computing and easier to interpret than using gradient descent and backpropagation to train feed forward networks, especially with fully connected layers involved. This motivates our new approach aiding the training of neural networks through first obtaining a PWL fit using spline-based method, then converting the model into a neural network with required hyperparameters and continue the training process in the standard way.

\subsection{Building a ReLU Neural Network From a MARS Model}
Here we present our method constructing and training a single-output ReLU neural network exploiting the MARS algorithm. We will soon show that it is possible to generalize this process to a much larger class of networks as well as other spline methods. We mainly focus on this specific case because ReLU is one of the most extensively used activation functions and the MARS algorithm is powerful on high-dimensional problems producing basis that can be directly applied on ReLU networks. The procedure consists of 4 steps:

\textbf{Step 1: Perform MARS on the dataset}

We assume the learning data have one-dimensional output for simplicity. There are hybrids of MARS that can be used to meet the need of multidimensional output \cite{friedman1991multivariate}, details of which will be discussed in the next part. As only first-order basis are added, the MARS algorithm here can be simplified. MARS usually does not need the whole data set to select basis, and empirical results suggest that separating out an exclusive part of data to train the MARS model, if applicable, can help reduce overfitting of the final model (see step 4).The produced spline model has the form:
\begin{equation}
    f(X) = \beta_{0} + \sum_{m=1}^{M} \beta_{m}h_{m}(X_{d_{m}}) 
\end{equation}
where M is the number of first-order basis, and $h_m(X)$ has the form $R(X_{d_{m}}-t_{m})$ or $R(t_{m}-X_{d_{m}})$.

\textbf{Step 2: Construct an Equivalent FC ReLU Networks}

In this step, we construct a three-layer, FC ReLU network that represents exactly the same function as the MARS model from Step 1. 
In the converted network, as illustrated in Figure 1, the number of units (width) in the input layer is the same as input dimension, and the width of the hidden layer is the number of basis, and the width of the last layer is output dimension.

\begin{figure}[H]
\centering
\includegraphics[width = 2in]{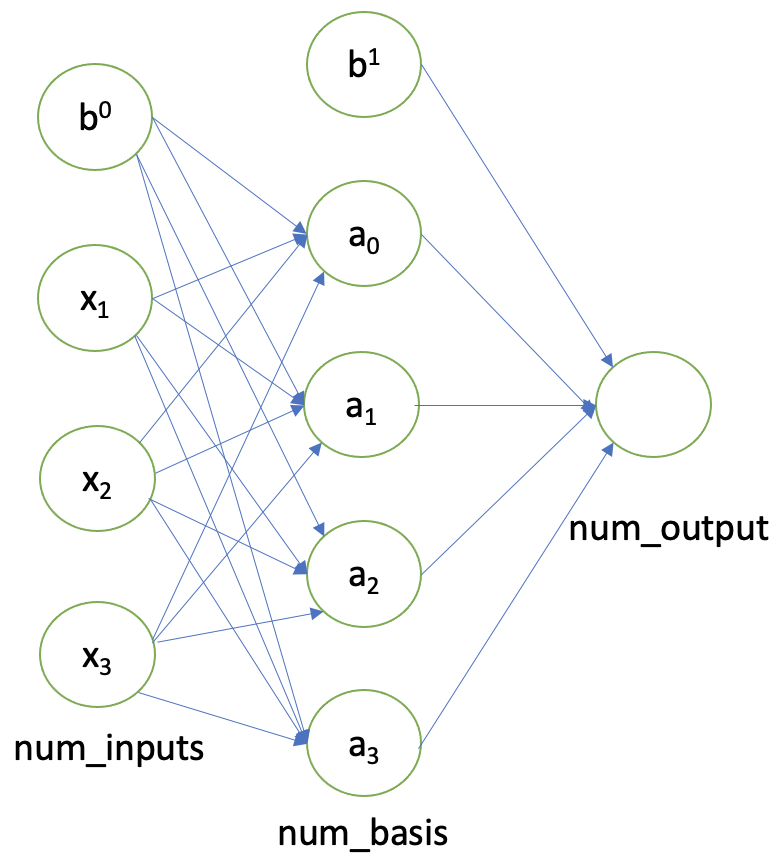}
\caption{Structure of Converted Layers}
\label{fig:universe}
\end{figure}

Parameters of the network is calculated from the linear spline model in the form of (7). For the weight matrix ${W^{(1)}}$ of the hidden layer, ${w^{(1)}_{i, j}}$ equals 1 if basis $h_i(X)$ is of the form $R(X_{j}-t_{i})$, -1 if $h_i(X)$ is of the form $R(t_{i}-X_{j})$ and 0 otherwise. The bias vector of the hidden layer is set to be ${(t_{1}, \cdots, t_{M})^{T}}$. The weight matrix (vector) of the output layer is calculated as ${(\beta_{1}, \cdots, \beta_{M})^{T}}$ (here we assume the output dimension to be 1, as mentioned), and the bias of the output layer is equal to the intercept ${\beta_{0}}$. Shown in Figure 2 is a simple example of the described conversion.

\begin{figure}[H]
\centering
\includegraphics[width = 3in]{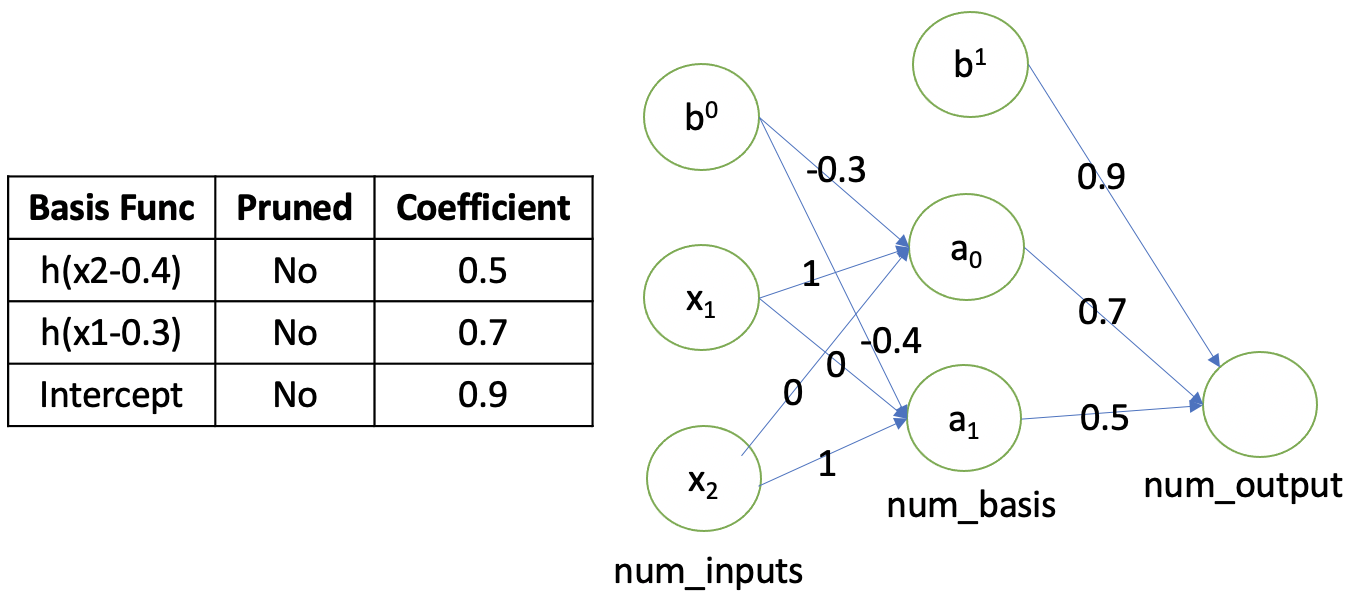}
\caption{One Example of Construction}
\label{fig:universe}
\end{figure}

\textbf{Step 3: Reshape the Network}

The converted network resulting from Step 2 may have an undesirable shape. The number of basis function produced by MARS is determined by minimization of GCV and is usually small comparing to the width of neural network in practical use. We therefore extend the raw network through the following two operation:
\begin{itemize}
    \setlength{\itemsep}{0pt}
    \setlength{\parsep}{0pt}
    \setlength{\parskip}{0pt}
    \item Add neuron units to hidden layers and set weights and bias of their inputs all zero, which adds to the width of the network and preserves the function represented by the network.
    \item Insert a new hidden layer (with same width as its previous layer), and set the weight matrix of its input as an identity matrix and set the bias vector zero. This increases the depth of the network by one, but the function represented by the network remains unchanged.
\end{itemize}

These two operations together reshape the network from Step 2 to a network with arbitrary larger width and depth.

\textbf{Step 4: Proceed with Gradient Descent Optimization Algorithms}

The conversion described above maps a linear spline model to a point in the neural networks' parameter space and we use this position as a start point for further training with gradient descent optimization algorithms to build a more elaborate model. The MARS model in this method is set to be additive and sometimes cannot depict the full picture of the input data pattern. By continuing with training, the expressiveness of the model is released from limitations of the original additive model.

\subsection{Extensions on PWL Function to FC Networks}
We now discuss several possible extensions to the method presented above. So far we mainly discuss building FC neural networks with our method, but it is applicable to the FC layers as part of a larger network that comprises other types of layers. For example, deep convolutional neural networks usually consist of convolutional and pooling layers with FC layers at head. With convolutional and pooling layers initialized by pretrained networks, which is common in visual tasks, the followed FC layers can be initialized from spline models applying on the CNN codes (output from previous layers).

To further support such scenarios, we consider the construction of multi-output networks as mentioned earlier. In this case, we can use a variation of MARS, PolyMARS \cite{stone1997} for multi-output regression problems and use its polychotomous logistic likelihood version\footnote{An implementation in \textbf{R} is known as {\it Polyclass}.} to deal with multi-class classification tasks. 

It is further possible to use another spline method as long as it outputs a CPWL fit to the data set. The construction of neural network needs to be generalized correspondingly. This leads to a general algorithm converting a PWL function into a feed-forward FC neural network that performs the same function. For this, it is convenient to exploit a previous study on the representation of PWL functions. The so-called {\it lattice PWL functions } proposed by Tarela et al. \cite{Tarela-PWL} have the following form:
\begin{equation}
    \min_{1 \leq i \leq M} \max_{j \in s_i} \ [1 \ x^{T}]\theta(j)
\end{equation}

Using a representation of any PWL function by such a lattice form \cite{TARELA199917}, it can be converted to a nested absolute-value functions and further FC layers. For each maximum function take the following representation:
\begin{equation}
    \max(x, y) = \frac{x+y}{2} + \max(\frac{x-y}{2}, 0) + \max(\frac{y-x}{2}, 0)
\end{equation}
and similarly for the minimum function take
\begin{equation}
    \min(x, y) = \frac{x+y}{2} - \max(\frac{x-y}{2}, 0) - \max(\frac{y-x}{2}, 0)
\end{equation}

This is taken further to neurons that express the minimum and maximum functions. Then by applying the conversion inductively on first the max operator and then the min operator, we can obtain a ReLU neural network with maximum depth ${log_{2}(M) + log_{2}(\max\{ |s_{i}|\})}$ that represents the same function as the PWL function. Note that there are other ways to do this conversion if we represent the PWL function in a different form, and one example is outlined in \cite{Arora-Rectified-Linear-Units}. Though we focus on neural networks with ReLU as its activation function, the algorithm can be further extended to apply on networks with some other CPWL activation functions such as Leaky ReLU \cite{maas2013rectifier} as long as it can represent the absolute function or the max operator.

\section{Theoretical Analysis}
In this section, we demonstrate three improvements our method makes to the training of FC layers, the time cost, the convergence performance, and the interpretability. As mentioned earlier, we will mainly consider first-order MARS as a detailed example of spline method for analysis.

\subsection{Time Cost}
Considering the time cost, our method can be viewed as replacing part of the standard training process with building a linear spline model plus constructing an equivalent neural network. The time consumption of construction for FC layers is negligible comparing to that of training processes. Therefore, we compare the time cost of two methods simply by calculating the time complexity of first-order MARS and FC networks. Let the input dimension be $d$ and the number of inputs be $N$. The time complexity of our first-order MARS method\footnote{The proof is provided in the appendix.} is
\begin{equation}
    \Theta(N\cdot (d\cdot M_{max}^3+M_{max}^4))
\end{equation}

In the FC network, if the width of the layers are $a_1,a_2,a_3, ..., a_l$, respectively, the input dimension is $a_1=d$, the output dimension is $a_l$, and the depth of the network is $l$, Previous work \cite{NNcomplexity} has shown that the total time complexity of the FC network in $m$ epochs is
\begin{equation}
    \Theta(m\cdot N\sum_{i=1}^{l-1}a_{i}a_{i+1})
\end{equation}

And in most cases, as $a_1=d$, the minimum value of $a_2$ is the number of spline basis functions,
\begin{equation}
    \Theta(\sum_{i=1}^{l-1}a_{i}a_{i+1})>\Theta(d\cdot M_{max}+M_{max}^2)
\end{equation}

In general, we consider MARS outperforms FC networks in efficiency for the following reasons: 

\begin{itemize}
    \item Although the time complexity of both first-order MARS and the neural network is linear in $N$, the factor $d\cdot M_{max}^3+M_{max}^4$ is limited, whereas the factor $\sum_{i=1}^{l-1}a_{i}a_{i+1}$ can grow extremely large as the neural network becomes deep and wide.
    \item The time complexity for neural network accounts for $m$ epochs, but the time complexity for MARS above is the upper limit of time cost for obtaining a MARS fit. In order to train a neural network with similar approximate accuracy as MARS, we usually need far more epochs than $M_{max}^2$. In this way, $\Theta(m\cdot N\sum_{i=1}^{l-1}a_{i}a_{i+1})>\Theta(M_{max}^2\cdot N\sum_{i=1}^{l-1}a_{i}a_{i+1})>\Theta(M_{max}^2\cdot N\cdot (d\cdot M_{max}+M_{max}^2))=\Theta(N\cdot (d\cdot M_{max}^3+M_{max}^4))$. Therefore, the overall time cost of neural network will be much lower.
    \item MARS is exploited to give a rough fitting to the dataset and the accurate approximation is performed further by neural network. Therefore, it is applicable to reduce the sample points needed to fit MARS, and the reduced $N$ will result in a more lower time cost for MARS.
    \item Our experimental results also show that the time consumption for MARS is much lower than the one for neural networks. To be more precise, the time cost of MARS is less than the time to train neural network for two epochs in our experiment.
\end{itemize}

Above all, we claim that our method has a low time cost and is meaningful in practical applications.

\subsection{Convergence Performance}
The analysis in this part focuses on applying our method on ReLU neural networks. For a neural network with randomly initialized parameters, its initial error is high and when the size of network increases, the exponentially growing number of saddle points surrounded by high error plateaus would significantly slow down training \cite{Identifying-and-Attacking}.

In contrast, models produced by spline methods such as MARS are able to reach a considerably low error in a large class of problems. So the network constructed from our method, as it is an equivalent operator to the linear spline model, will have its parameters located close to a minimum at the start of training. This neighborhood minimum, either local or global, is guaranteed to be no worse than the starting point found by the spline method regarding to the training loss. In experiments, the low initial test error or loss and the small parameter shift after training confirm our analysis empirically. In this sense, our method helps avoid unnecessary steps that are used for gradient descent methods to flee from saddle points (or reduce the risk of getting stuck into these points) and ensures that the training would converge to a overall good minimum point.

\subsection{Interpretability}
In our approach the resulting spline functions have the form $f(x) = \beta_{0}+\sum_{m=1}^{M} \beta_{i}h_{m}(X)$, where the $h_{m}(X)$ are basis functions of MARS related to a particular input dimension. The entry ${X_{j}}$ of input in $h_{m}(X)$ contributes to the prediction of the output data, and a larger $\beta_{i}$ indicates a larger importance of corresponding input dimension. On the other hand, our method encourages a sparse solution, especially when regularization methods such as $L_{1}$ regularization or dropout \cite{dropout} instead of $L_{2}$ regularization is applied during the training process. Such sparse networks are generally easier to interpret and have advantages on computational efficiency \cite{2015Neural}. In this sense, we can better explain model functionality.

Furthermore, after fitting MARS, the selected functions $h_{m}(X)$ comprise several reflection pairs $R(\pm(X_{j}-t))$, where the {\it knot value} $t$ is one of the sample points. By looking into which knots are chosen by MARS, we can determine which sample points are important to the model. This will also enhance the model transparency. Considering the improvement on explaining model functionality and model transparency, the interpretability of the converted neural network is generally increased \cite{Chakraborty-interpretability}. 

\section{Experiments}
\subsection{Datasets}
We employ two datasets {\it Abalone\/} and {\it Wine Quality\/} to evaluate our method. The data in {\it Abalone} is used for predicting the age of abalone from physical measurements. The data is available from UCI Machine Learning Repository, and consists of 4177 samples with eight numerical attributes and one categorical attribute. We divide the dataset into a training test dataset with a ratio of 7:3.  

The {\it Wine Quality\/} dataset is also obtained from UCI Machine Learning Repository; it concerns the modeling of wine quality based on physicochemical tests. The dataset consists of 4898 samples with eleven numerical attributes and one categorical attribute. It is also split with a ratio of 7:3.  Both datasets are normalized and randomly shuffled before training and testing to avoid unbalance. 
\subsection{Implementation}
For the purpose of obtaining a piecewise-linear spline which is transformed into fully-connected layers of deep neural network, we apply first-order MARS\footnote{Use Python package {\it sklearn-contrib-py-earth}.}. The process of transformation has already been depicted in the previous methodology part, section 3.2. \\
\subsection{Results}
For the {\it Abalone} dataset, our method significantly reduces the initial MSE loss of the converted network, and sets the network to continue training. Results show than our method saves time of a large fraction of epochs. A comparison of the losses of converted network and the network with randomly initialized parameters and identical shape is shown in Figure 3.
\begin{figure}[H]
\centering
\includegraphics[width = 3in]{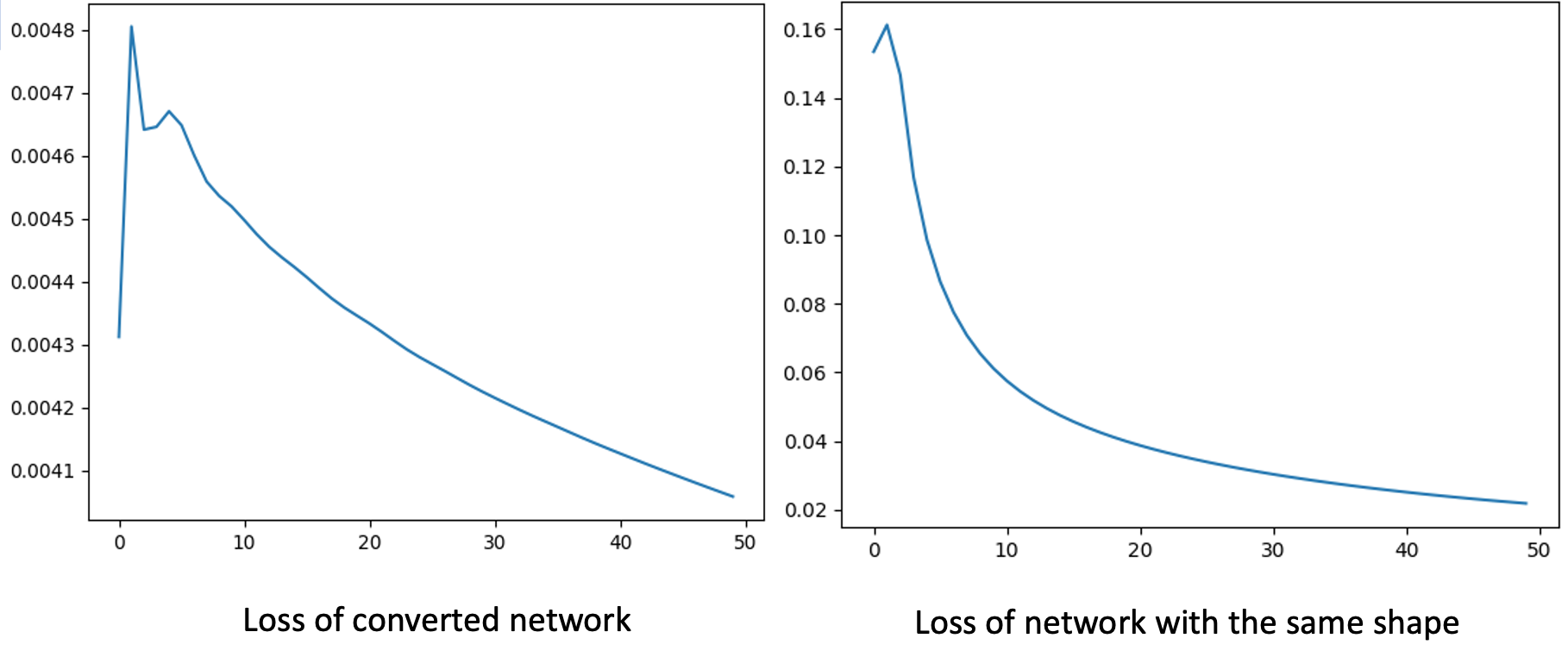}
\caption{Losses Comparison}
\label{fig:universe}
\end{figure}
Furthermore, we also tested the error on divided test sets, and found that our result is reasonable and did not over-fit the dataset. We keep track of the error before and after training respectively for 50/300/500 epochs. A detailed comparison of error is shown in Table 1, 2 and 3. From the experiment, we investigated that our method not only promoted the training, but also lowered the convergence error.
\begin{table}[H]
\caption{Error on Testsets After Training 50 Epochs}
\begin{tabular}{|l|l|l|}
\hline
epoch = 50           & Before Training & After Training  \\ \hline
Converted Network    & 0.0048424004  & 0.00438868788 \\ \hline
Randomly Initialized & 0.1495334970  & 0.00615985973 \\ \hline
\end{tabular}
\end{table}

\begin{table}[H]
\caption{Error on Testsets After Training 300 Epochs}
\begin{tabular}{|l|l|l|}
\hline
epoch = 300           & Before Training     & After Training        \\ \hline
Converted Network    & 0.0048424004 & 0.00202874085   \\ \hline
Randomly Initialized & 0.1128622549 & 0.00341153305 \\ \hline
\end{tabular}
\end{table}

\begin{table}[H]
\caption{Error on Testsets After Training 500 Epochs}
\begin{tabular}{|l|l|l|}
\hline
epoch = 500           & Before Training & After Training \\ \hline
Converted Network    & 0.0048424004   & 0.00140022880 \\ \hline
Randomly Initialized & 0.1493096528   & 0.00483068342 \\ \hline
\end{tabular}
\end{table}
We further compared the time cost of our spline method with a pure neural network, where both time measurements were completed with only CPU used\footnote{using Mac with 2.2 GHz Quad-Core Intel Core i7-4770HQ Processor, 16GB RAM}. As data in Table 4 shows, the time it takes for fitting splines is even shorter than the time it takes for two epochs, but it strikingly reduces the training loss and finds a good starting point for the subsequent training. 
\begin{table}[H]
\caption{Time Usage When Training 100 Epochs}
\begin{tabular}{|l|l|l|}
\hline
Time                                                                  & \begin{tabular}[c]{@{}l@{}}Total Time \\ (100 epochs)\end{tabular} & Time Per Epoch         \\ \hline
Spline Fitting                                                        & 0.00822214502                                                      & \multicolumn{1}{c|}{/} \\ \hline
\begin{tabular}[c]{@{}l@{}}Converted Network \\ Training\end{tabular} & 5.21306848002                                                      & 0.0521306848002       \\ \hline
\begin{tabular}[c]{@{}l@{}}Control Network \\ Training\end{tabular}   & 5.10581314598                                                      & 0.0510581314598       \\ \hline
\end{tabular}
\end{table}
In addition, in order to measure how well the spline model fits compared with the final convergence, we compared the parameters of the converted network before the training and after it.

The following two matrices represent some weight parameters in the hidden layers before and after training. Note that the left matrix is the network directly converted from spline without further training, and the right matrix is the parameters after training. \\

\begin{equation}
\begin{pmatrix}
0 & 0 & 0\\
0 & 0 & 0\\
0 & 0 & 0\\
0 & 0 & 0\\
1 & 0 & 0\\
0 & 0 & 0\\
0 & 1 & 0\\
0 & 0 & 1\\
0 & 0 & 0\\
0 & 0 & 0
\end{pmatrix}
\begin{pmatrix}
0.0075 & 0.0076 & 0.0050\\
-0.0011 &  -0.0012 & 0.0022\\
0.0208 & 0.0208 & -0.0024\\
0.0001 & 0.0002 & 0.0129\\
0.9967 & -0.0032 & 0.0141\\
0.0015 & 0.0017 & 0.0144\\
-0.0155 & 0.9845 & 0.0173\\
0.0026 & 0.0026 & 1.001\\
-0.0171 & -0.0171 & 0.0186\\
-0.0214 & -0.0214 & 0.02232
\end{pmatrix}
\end{equation}

Comparing the two matrices representing the weight parameters we find that the change is relatively small, so we claim that our initialization point is already quite close to the actual minima, which make it a good starting point.

We also tested our method on the {\it Wine Quality\/} dataset. The comparison of MSE losses is shown in Figure 4. Similar to the loss results for {\it Abalone} dataset, the initial loss of converted network is close to the convergence loss of the network with identical shape. 
\begin{figure}[H]
\centering
\includegraphics[width = 3.5in]{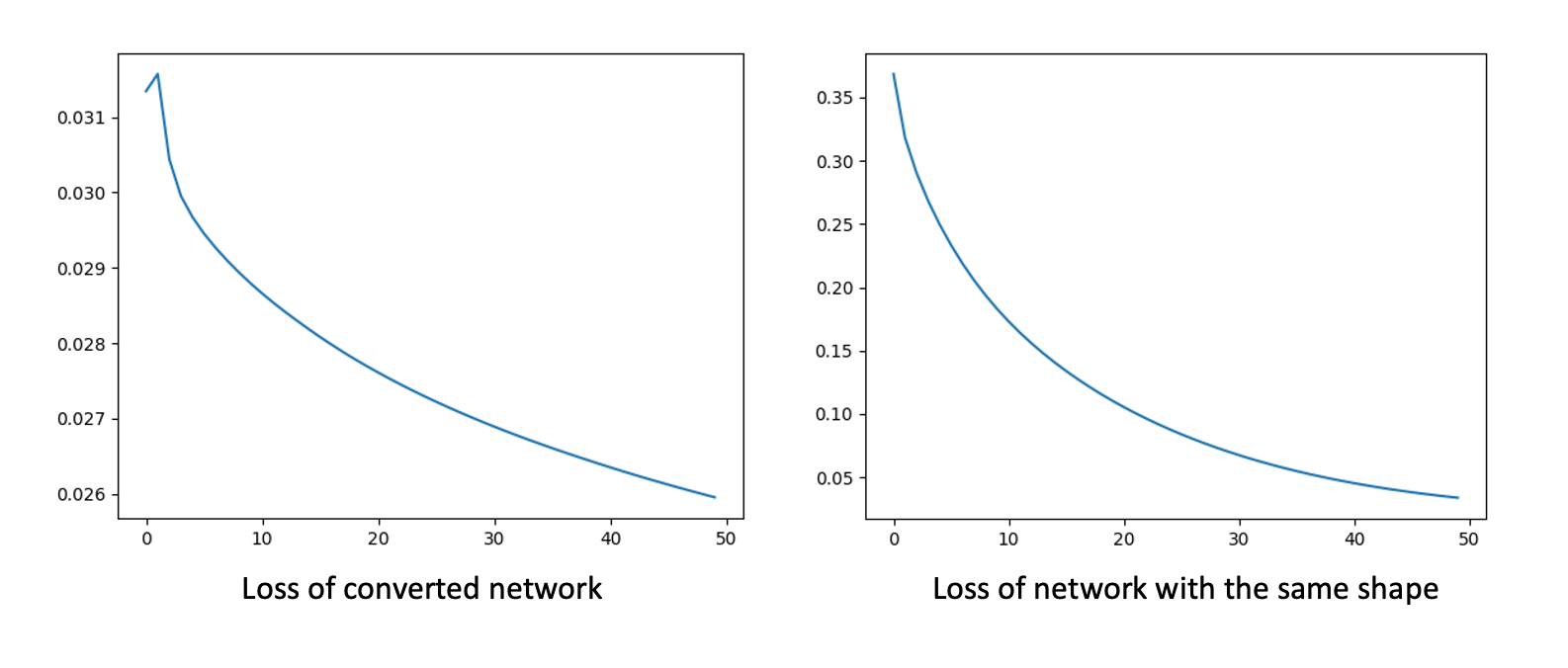}
\caption{Losses Comparison}
\end{figure}
Concerning the error tested on the testset, the result also shows that our method avoided the problem of overfitting and lowered the error. 
\begin{table}[H]
\caption{Error on Testsets After Training 50 Epochs}
\begin{tabular}{|l|l|l|}
\hline
Time                 & Before Training & After Training \\ \hline
Converted Network    & 0.04454497663   & 0.036338012  \\ \hline
Randomly Initialized & 0.54242669885   & 0.044301751  \\ \hline
\end{tabular}
\end{table}

Similar to what we have done on {\it Abalone} dataset, we also measure the time cost with CPU only\footnote{using Mac with 2.2 GHz Quad-Core Intel Core i7-4770HQ Processor, 16GB RAM}. Though the time cost is larger since the dimension and number of samples is larger on {\it Wine Quality} dataset, the time cost for spline fitting is still slightly smaller than time for two epochs. 
\begin{table}[H]
\caption{Time Usage When Training 100 Epochs}
\begin{tabular}{|l|l|l|}
\hline
Time                                                                  & \begin{tabular}[c]{@{}l@{}}Total Time \\ (100 epochs)\end{tabular} & Time Per Epoch         \\ \hline
Spline Fitting                                                        & 0.193164607975                                                      & \multicolumn{1}{c|}{/} \\ \hline
\begin{tabular}[c]{@{}l@{}}Converted Network \\ Training\end{tabular} & 12.46365711896                                                      & 0.1246365711896       \\ \hline
\begin{tabular}[c]{@{}l@{}}Control Network \\ Training\end{tabular}   & 12.43784837599                                                      & 0.1243784837599       \\ \hline
\end{tabular}
\end{table}

\section{Conclusion}
In this paper, we studied the relationship between linear splines and deep neural networks, and proposed a spline-based algorithm that eases the difficulty of training FC layers. To be more specific, we first obtain a CPWL fit and convert it to FC layers of an ANN. Then, through the comparison of time cost, convergence performance and interpretability we provide evidence that our method helps to increase the efficiency and accuracy of training as well as enhance the interpretability of ANN. We further performed experiments on the {\it Abalone} and {\it Wine Quality} datasets to demonstrated that our method improves the training performance significantly and to support our theoretical analysis. In our future work, we intend to further explore the application of our method in visual representation transfer  and classification problems, as well as investigate the method that converts our fitting of splines to a larger class of neural networks.

\newpage

\bibliographystyle{named}
\bibliography{ijcai21}

\appendix
\section{Time Complexity of First-Order MARS}

In the first-order MARS model, if there are $M$ basis functions, the time complexity of minimizing $GCV(M)$ is $\Theta(N\cdot M^2)$, as minimizing $GCV(M)$ is done analogously to linear regression.

In the forward phase, we need to obtain $M_{max}$ basis functions, and for each basis function, there are $d$ different choices for the input dimension and there are $N$ different choices for the knot value $t$, as the knot value is chosen from the values of the related dimension of the inputs. Assuming there are already $M-1$ basis functions, the time complexity of determining the new basis function is $\Theta(d\cdot N\cdot N\cdot (M-1+1)^2)=\Theta(d\cdot N^2\cdot M^2)$. As the procedure stops when the number of basis functions reach the maximum $M_{max}$, the total time complexity of the forward step is $\Theta(M_{max}\cdot d\cdot N^2\cdot M_{max}^2)=\Theta(d\cdot N^2\cdot M_{max}^3)$. In addition, in previous researches it was shown that with techniques avoiding repeated calculations, the time complexity can be optimized to $\Theta(d\cdot N\cdot M_{max}^3)$ \cite{hastie01statisticallearning,friedman1991multivariate}.

In the backward phase, we obtain $M_{max}-1$ models, and each model requires at most $M_{max}-1$ linear regressions, so the total time complexity is $\Theta((M_{max}-1)\cdot (M_{max}-1)\cdot (N\cdot M_{max}^2))=\Theta(N\cdot M_{max}^4)$. 

Therefore, the time complexity of first-order MARS is
\begin{equation}
    \Theta(N\cdot (d\cdot M_{max}^3+M_{max}^4))
\end{equation}

\end{document}